\documentclass{llncs}

\usepackage{graphicx}
\usepackage[utf8]{inputenc}

\newcommand{\ie}{{\it i.e.}}
\newcommand{\eg}{{\it e.g.}}

\newcommand{\etc}{{\it etc.}}

\begin{document}

\title{Cooperative Interface for a Swarm of UAVs}
\author{Sylvie Saget\inst{1,2} \and Fran\c{c}ois Legras\inst{1,2} \and Gilles Coppin\inst{1,2}}
\institute{Institut TELECOM; TELECOM Bretagne; UMR CNRS 3192 Lab-STICC, France
\and Universit\'e Europ\'eenne de Bretagne, France}

\maketitle

\begin{abstract}
	After presenting the broad context of authority sharing, we  outline how introducing more natural interaction in the design of the ground operator interface of UV systems should help in allowing a single operator to manage the complexity of his/her task. Introducing new modalities is one one of the \emph{means} in the realization of our vision of next-generation GOI. A more fundamental aspect resides in the \emph{interaction manager} which should help balance the workload of the operator between mission and interaction, notably by applying a multi-strategy approach to generation and interpretation.

	We intend to apply these principles to the context of the \textsc{Smaart} prototype, and in this perspective, we  illustrate how to characterize the workload associated with a particular operational situation.
\end{abstract}

\section{Introduction}

Unmanned Vehicle (UV) Systems will considerably evolve within the next two decades. Actually, in the current generation of UV Systems, several ground operators operate a single vehicle with limited autonomous capabilities. Whereas, in the next generation of UV Systems, a ground operator will have to \textit{supervise} a \textit{system} of several cooperating vehicles performing a joint mission, i.e. a Multi-Agent System (MAS) \cite{Joh03,LC07}. In order to enable mission control, vehicles autonomy will  increase \cite{DW01} and will require new forms of Human-system interaction.

In this context, we have developed a prototype multi-UV ground control station (\textsc{Smaart}) that allows an operator to supervise the surveillance of a simulated strategic airbase by a swarm of rotary-wing UVs \cite{legras2008experiments}. In this system, the autonomous behavior of the UVs is generated by the means of a dual digital pheromone algorithm (bio-inspired approach), and although we obtained interesting results, the operator-system interaction is rather basic: place beacons in the environment, dispatch UVs towards these beacons, select mode of information display, \etc{} Despite being technically an \emph{authority sharing} system, \textsc{Smaart} is only the first step in the development of a full-fledged authority sharing control system for swarms of UVs.
Two main challenges have to be faced in order to design such an efficient and realistic UV Systems:
\begin{enumerate}
	\item obviating the negative side effects of automation: workload mitigation, loss of situation awareness, complacency, skill degradation, etc \cite{PSW00};
	\item decreasing the cognitive load induced for the ground operator. Operating UV systems is highly complex. Obviously, shifting to UV Systems with several vehicles will makes mission and vehicles control more complex \cite{CBMM07}. In addition, even though increasing vehicles' autonomy aims at decreasing the cognitive load induced by mission control for ground operators, workload mitigation may lead to even higher workload \cite{PSW00,CBMM07}.
\end{enumerate}

While using an UV System's interface, the ground operator is at least engaged within two activities: mission control and interaction. However, all in all, the great majority of study focus on mission control. In this paper, we claim that interaction design must be considered as a field of research by itself. In this perspective, as  \cite{WDCB05,GNBWSC02} we claim that there is a need to enhance the naturalness of ground operator interface rather than only improving mission realization and control. But as soon as an interface provides “natural” input and/or output devices, non-understandings and misunderstandings may occur. Then, in order to design efficient UV Systems, the problem of  robustness of the interaction has to be handled. 

Based on recent advances in Pragmatics and in Human-Computer Interface \cite{Cla96,Tra94,PG04,SG06}, we present a collaborative view of interaction dedicated to UV Systems. Considering interaction as a collaborative activity while designing an interface enhances its robustness \cite{SLC08}, opens the door to  managing the global workload of the operator through a balance effect between mission load and interaction load \cite{MGKH01}.

\section{Perspective on Authority Sharing}

\begin{figure}[ht]
	\centering
		\includegraphics[width=.3\textwidth]{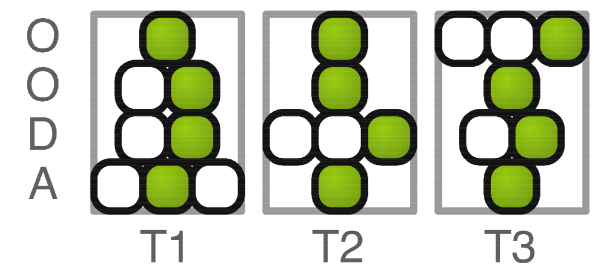}
	\caption{Representation of the operating modes for three tasks (T1, T2, T3).}
	\label{figOpMod}
\end{figure}

This section presents the backdrop of our research on authority sharing for unmanned systems control. The basic requirement for an authority sharing system is to provide several distinct operating modes to accomplish a given task or function. Fig.~\ref{figOpMod} illustrates how we represent the different operating modes for three tasks, decomposed along John Boyd's OODA loop \cite{ullman2007oo-oo-oo}. In this example, the system has only one operating mode for the \emph{Observe} stage of T1, but two alternative modes for \emph{Orient} and \emph{Decide}, and three modes for \emph{Act}.

Each operating mode gives specific authority and responsibilities to the operator and the system. For example, the \emph{Act} stage of the landing task of an UV could provide three modes: (a) full manual \ie{} the operator tele-operates the UV; (b) full auto \ie{} the vehicle lands automatically with no possible intervention of the operator; and (c) auto with veto, where the operator can disconnect the auto-pilot and handle the landing.

\begin{figure}[t]
\begin{center}
\includegraphics[width=\columnwidth]{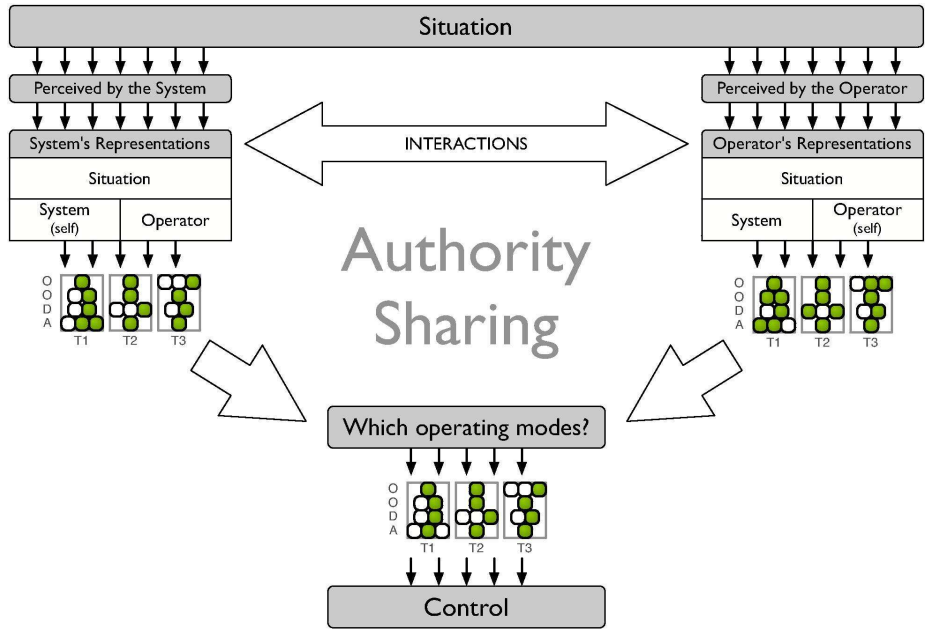}
\caption{Authority sharing concepts in single system--single operator interactions.}
\label{figPA}
\end{center}
\end{figure}

As we see on Fig.~\ref{figPA}, in order to control the system at time $t$, an operating mode must be activated for each task and stage in OODA (filled cells on Fig.~\ref{figOpMod}). Therefore, the main question is ``which operating mode has to be selected at time $t$?'' This corresponds to a second level of authority sharing \ie{} the authority to assign responsibilities (select operating modes) and must be implemented as some sort of decision process. Indeed, in a full-fledged authority-sharing system, the operators and the machines could have different preferences concerning which operating modes should be selected.

We can consider that the operators and the system's internal representations can be broadly decomposed along the same three categories:
\begin{itemize}
	\item models and representations of the situation: his corresponds to the representation of objects in the system's environment, knowledge about its laws, properties, \etc{}: this can be seen as a ``world model'';  
	\item models and representations of the system: his corresponds to the current state of the system, its known capabilities, predictions about its evolution, \etc;
	\item models and representations of the operator: similarly this represents the state of the operator, his/her abilities, performances, \etc
\end{itemize}

If one combines these categories with the two kind of actors  (machine and Humans), we obtain six distinct fields of research relevant to the development of authority sharing systems. Without being exhaustive, one can identify:
\begin{itemize}
	\item work on human situational awareness (representation of the situation on the operator's side);
	\item work on trust in automation (representation of the system on the operator's side);
	\item cognitive and physiological modeling (representation of the operator on the system's side).
\end{itemize}

System's and operator's representations are not only fed by observation of the situation (as illustrated on Fig.~\ref{figPA}), but also through interaction between Human and machine. Man-machine interaction can happen on several media (from classic mouse \& keyboard, joystick to advanced haptic interfaces or dialogue) but whatever the chosen media, it should tend to facilitate the convergence between the respective representations of Humans and machines involved. Again, we consider here another field of research by itself. And one can note that an efficient interaction will decrease the difficulty of the final decision-making process as converging representations on the Human and machine's sides lead to converging preferences on which operating modes to select.

\section{Naturalness}

In the current generation of UV Systems, Ground Operator Interface (GOI) is a traditional Graphical User Interface, such as \cite{Kim93}. These are based on input/output modalities such as drop-down menu or push button, with a constrained interaction language, providing quantitative spatial information and interaction, \etc{} This interaction language is similar to the low-level command language for vehicles, with quantitative spatial information for example. Thus, GOI naturalness is poor. However, current works aims at enhancing the naturalness of interface \cite{CHPJ06}. That is integrating a less-constrained interaction, at least a single natural modality as input \cite{MGKH01,GNBWSC02} (such as gesture, spoken or written language) or output \cite{GNBWSC02} (such as speech, haptic display), multi-modality, flexible interaction \cite{LC07}, providing qualitative spatial information and interaction \etc

As soon as naturalness is introduced within the GOI, non-understandings -- due to vagueness, ambiguity or underspecification -- have to be managed. Interactive management of non-understandings follows from the collaborative nature of interaction (grounding) and requires a new GOI component: an interaction manager \cite{SLC08}. Such an approach has been used within The WITAS project \cite{LGCP02} as well as within the GeoDialogue Project \cite{CWME03}, which relates to Geographical Information Systems.

Enhancing GOI naturalness has various benefits for UV Systems. First, a more natural interaction between an interface and its user generally enhances the efficiency of interaction \ie{} it reduces the cognitive load induced by the interaction for the user as well as  interaction time. For example, a data entry function based on vocal keyword recognition may require a single vocal utterance in the next generation of GOI, while it may require over twenty separate manual actions in the current generation of GOI.

Moreover, natural display modality (typically, haptic display) also aims at making up for the “sensory isolation” of the ground operator \cite{GNBWSC02}. Operator's sensory isolation is due to the fact that he is generally not collocated in the same physical space than vehicles \cite{End95,SMW04}. This leads to lack of situation awareness, among others negative effects.

Supervisory control of UV systems mainly involves spatial cognition and reference to vehicles, landmarks, waypoints, \etc{} To the extend that human beings performs these tasks using qualitative information and  interact through gesture and verbal communication, GOI naturalness should focus on spatial information and interaction. Nonetheless our goal is not to transform the GOI in a fully natural interface. But, as soon as naturalness is introduced within an interface, side effects have to be carefully considered. In particular, the GOI must contain new functionalities: being a semantic bridge between operators and vehicles, and handling of non-understandings. 

	\begin{figure}[ht!]
	\centering
		\includegraphics[width=\textwidth]{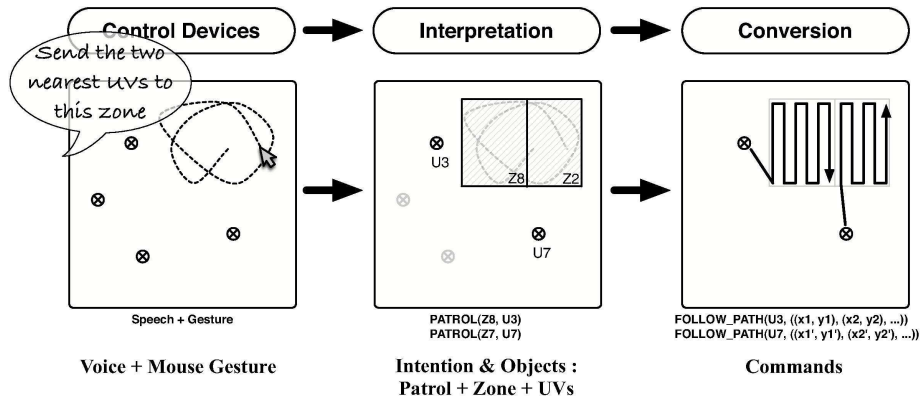}
	\caption{Ground operator interface: a semantic bridge.}
	\label{figTraduc}
\end{figure}

First, considering “natural” input device (\textit{i.e.} corresponding to a control command from the ground operator to a vehicle), there is a mismatch between the “natural” command provided by the operator and the “operational” command that a vehicle can accept. Then, the ground operator interface must be a \textit{semantic bridge}, that  converts the perceived message in a representation which is suitable for the addressee. That is to say that following the perception of an input on a control input device and following its interpretation, GOI also has to \textit{convert} the understood control command before transmitting it to the proper vehicle(s). As shown on Fig.~\ref{figTraduc}:
\begin{itemize}
	\item the front part the GOI has to detect input on a control input device and has to transmit the raw message to the interpretation module;
	\item the interpretation module identifies the intend meaning of the operator: the kind of control command and the intended objects, the two vehicles which are the recipients of the message and the intended pre-defined zone;
	\item the conversion module has to:
		\begin{enumerate}
			\item translate operator command into the language which is proper for the addressee;
			\item process the underspecified elements, such as path planning. The conversion module may lack necessary information, such as the first waypoint. In this case, a completion request has to be send to the ground operator in order to complete the command conversion.
		\end{enumerate}
\end{itemize}   

Second, as soon as an interface provides semi-constrained interaction, qualitative spatial interaction \cite{CWME03}, natural (multi-)modality \cite{Tra94}, then \emph{non-understandings} may occur. Non-understanding is commonly set apart misunderstanding. In a misunderstanding, the addressee succeeds in communicative act's interpretation, whereas in a non-understanding he fails. But, in a misunderstanding, addressee's interpretation is incorrect. For example, mishearing may lead to misunderstanding.

Handling non-understandings is necessary for the GOI as soon as an input control command cannot be transmitted to vehicles. Human beings handle non understandings by interactively refining their understanding until a point of intelligibility is reached. This process is called “grounding” \cite{Cla96}. In order to design grounding, interaction has to be viewed as a collaborative process and an interaction manager has to be integrated within the GOI. For more details, the interested reader may refer to \cite{SLC08}. There is a common misconception that non-understandings are  considered as “communicative errors” one may tend to avoid, as well as understanding refinement. Although, either for Human-Human Interaction \cite{MTNO90} or for Human-Computer Interaction \cite{MT03}, non-understandings and their management process present lots of advantages. For example, feedbacks are cue that enable interaction partners to be aware of the level of understanding of each other. Through interaction refinement each interaction partner maintains an accurate representation of the other and this enhances the efficiency of future interactions.

More generally, enhancing interaction design may lead to positive side effects enhancing mission control. If one takes the example of gesture, as explained in the previous section, enabling the ground operator to interact using gesture aims at making efficient references. 
Actually, gestures facilitate the maintenance of spatial representations in working memory \cite{WHKW01}. Therefore, gestures may contribute to maintaining the situation awareness that enables an efficient supervisory control by the operator.

\section{Balancing Mission \& Interaction}
		
While using an UV System's interface, the ground operator is at least engaged within two activities: mission control and interaction. This is the general case of all goal-oriented interaction (or dialogue):

\begin{quote}
\emph{Dialogues, therefore, divide into two planes of activity \cite{Cla96}. On one plane, people create dialogue in service of the basic joint activities they are engaged in making dinner, dealing with the emergency, operating the ship. On a second plane, they manage the dialogue itself -- deciding who speaks when, establishing that an utterance has been understood, etc. These two planes are not independent, for problems in the dialogue may have their source in the joint activity the dialogue is in service of, and vice versa. Still, in this view, basic joint activities are primary, and dialogue is created to manage them.} \cite{BC03}
\end{quote}

Interaction is defined by each dialog partner's goals to understand each other, \ie{} words to reach a certain degree of intelligibility, \textit{sufficient for the current purpose}. The crucial points here are that :
\begin{enumerate}
    \item perfect understanding is not required, the level of understanding required is directed by the basic activity  (\ie{} the mission) and the situational context (\eg{} time pressure);
    \item as ground operator's cognitive load is “divided” between the cognitive loads induced by each activity, the interaction's complexity must vary depending on the complexity involved by the mission, as defined by Mouloua and al. \cite{MGKH01}. For example, as time pressure rises, the cognitive load induced by the mission increases. The cognitive load required by the interaction should decrease in order to carry through the mission.
\end{enumerate}

In the perspective of adapting  adapt the grounding process to the specific case of the supervision of multiple UVs, one can note that a GOI is also an interaction support for a team, therefore similar to interfaces dedicated to Computer Supported Cooperative Work. However, the team include humans (ground operators) \textit{and machines} (vehicles). The GOI is also an interaction partner of ground operators. They interact for decision support tasks, non-understanding management and interface manipulation.

\begin{figure}[ht]
\begin{center}
\includegraphics[width=0.5\columnwidth]{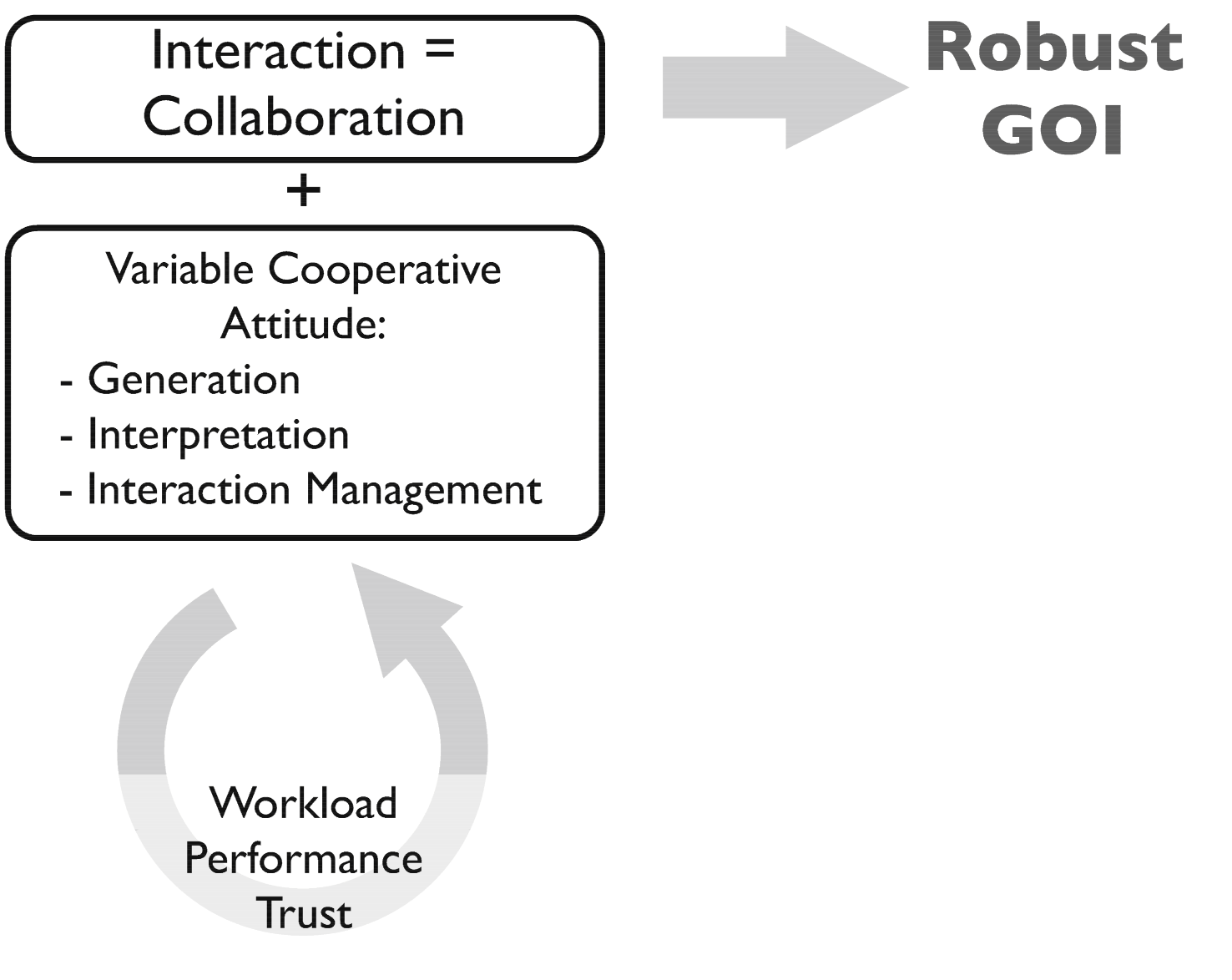}
\caption{Interaction as a collaborative and adaptive process.}
\label{figInteraction}
\end{center}
\end{figure}

In addition, our study aims at designing interaction in order to take  advantage of the mutual dependancy between mission control and interaction. More precisely, we propose adaptive interaction design in order to obtain the properties defined by Mouloua and al. \cite{MGKH01,SLC08}. The point is to obviate the negative side effects of automation, through balanced workload. We claim that adapting interaction design in regards human factors such as operator's workload, trust or performance must have positive side effects on these human factors, cf. Figure~\ref{figInteraction}.

The GOI we are developing will integrate an \emph{Interaction Manager} component, responsible for choosing adequate strategies for (1) \emph{generation} and (2) \emph{interpretation}:
		\begin{enumerate}
			\item for a given information or interaction to be potentially send to the operator, deciding whether to send it or not and choosing a modality and formulation;
			\item trying to understand interactions emanating from the operator, given the current context and grounding information.
		\end{enumerate}
Several redundant strategies will have to be developed for the various possible interactions permitted by the GOI. For a given interaction, some strategies will put the burden of interaction on the GOI (disambiguation, acknowledgement, \etc) while others will rest more “on the shoulders” of the operator. The rationale is that in situations of low “mission workload”, one is better off with giving more work the operator as it builds up the grounding all the while keeping him/her busy (fending off boredom and loss of attention). Conversely, in mission critical episodes, the GOI can assume a more active role in interaction and let the operator focus on the mission (and still be robust thanks to the grounding constructed earlier).

\section{Situation Cueing for the GOI of a Swarm of UVs}

In addition to designing different strategies for generation and interaction, we have to give to the interaction manager some means to evaluate the current state of the mission (and therefore make an estimation of the associated workload) in order to choose strategies. In this section, we illustrate how -- in the perspective of the extension of the \textsc{Smaart} system (see  \cite{legras2008experiments}  in this volume) -- we intend to discriminate four categories of \emph{mission states} and their associated workloads.  

\subsection{Description}

The \textsc{Smaart} system allows patrol and intercept operations for a dozen of rotary-wing UAVs supervised by a single operator.

\begin{figure}[t]
	\centering
		\includegraphics[width=.8\textwidth]{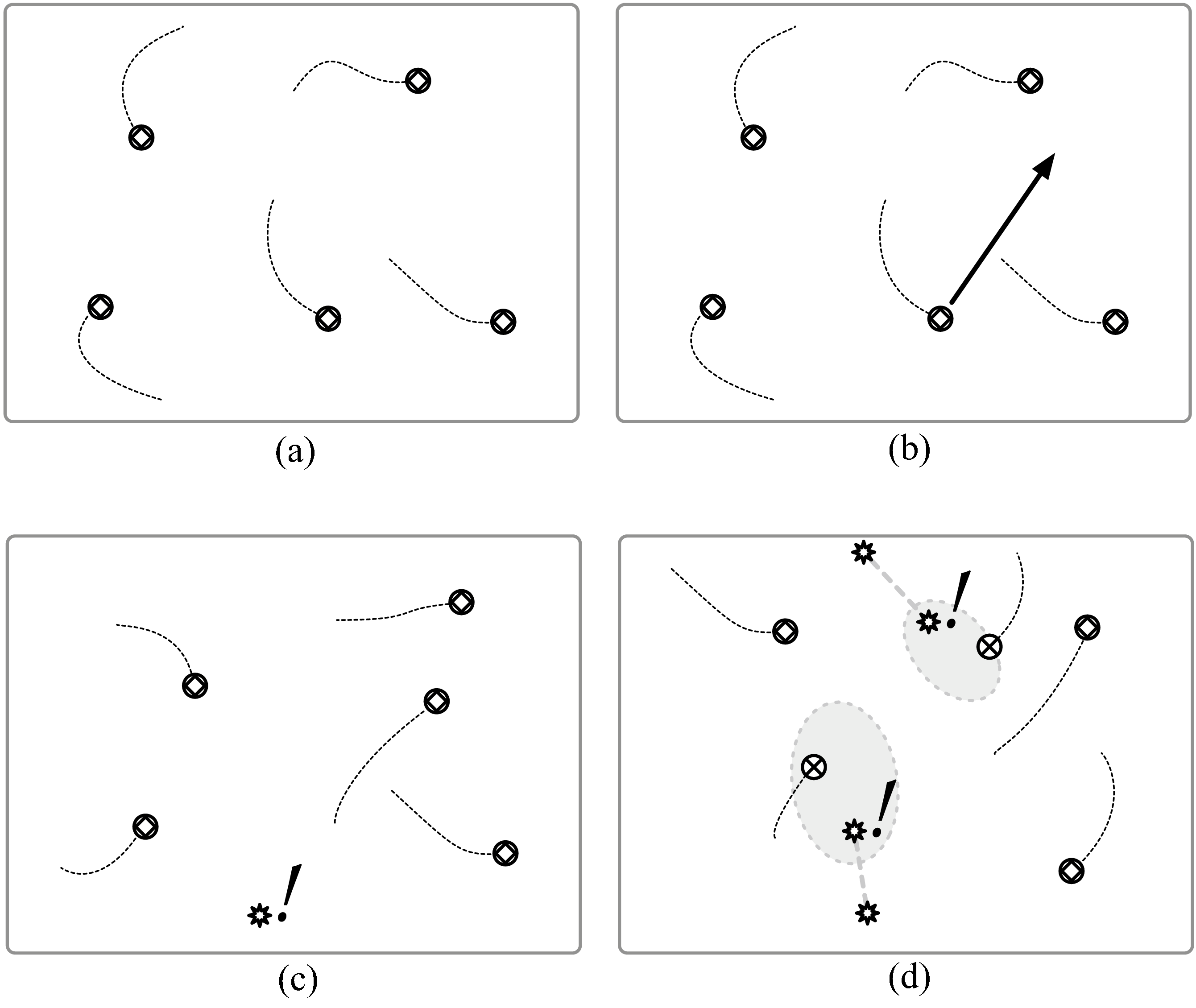}
	\caption{Illustration of levels of workload. Symbols {\protect\raisebox{-3pt}{\protect\includegraphics[height=4.4mm]{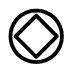}}} and {\protect\raisebox{-2pt}{\protect\includegraphics[height=4mm]{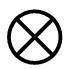}}} correspond to -- respectively -- patrolling and pursuing UVs. The {\protect\raisebox{-3pt}{\protect\includegraphics[height=5mm]{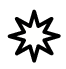}}}  symbols correspond to alarms. Thick dotted lines link alarms that are supposed to have been triggered by the same intruder, while greyed-out regions are search zones with adjustable parameters (center, direction and breadth). Recent alarms are indicated by an exclamation mark (!).}
	\label{figExamples}
\end{figure}

Subfigure~\ref{figExamples}a illustrates the lowest level of workload: \emph{Routine Patrol}. 
The UAVs perform their surveillance with a stable performance, every point on the airbase is scanned at an acceptable frequency.
In this context, the task of the operator is purely of a supervisory nature. The parameters of the display  are set to values known to be adequate for the current setup (number of UAVs, area, threat level, \etc{} see \cite{legras2008experiments} in this volume). The operator has a good appreciation of the reliability of the algorithm (good) and knows that anomalies are rare.

Due to its local nature, the algorithm that the UAVs use to perform their coordinated patrol can misbehave in some configurations.\footnote{For example, if for some reasons an ``islet'' appear in the digital pheromone space, the gradient-following UAVs will never reach it, and consequently, a dark spot will appear and worsen. This local processing on the part of the UAVs has some interesting properties, but also has the consequence that a modification in a part of the airbase (\eg{} creation of a no-fly zone) will take some time to be propagated to the rest of the environment though the digital pheromone.}
In such cases, the operator can detect the anomaly and take actions by dispatching some UAVs manually to compensate for the anomaly. In this context he/she has to closely supervise the execution of these actions, judge their effectiveness, all the while continuing his/her global supervising activity of the patrol on the whole airbase. One can detect such a workload level (\emph{Patrol with Anomaly}) by the action of the operator on an UAV (Subfigure~\ref{figExamples}b).

The two next workload levels are characterized by the presence of alarms. The number of alarms in recent time allows to distinguish low threat \emph{Alarm} (possible false alarm, Subfigure~\ref{figExamples}c) from emergency situation (multiple alarms, coordinated \emph{Intrusion}, Subfigure~\ref{figExamples}d). In this last situation, the general surveillance of the airbase is largely jeopardized, as (1) many UAVs are used to pursue the intruders in specific regions, therefore depleting the patrolling vehicles. And, (2) the attention of the operator is largely focused on the intrusions. 

\subsection{Characterization}

Table~\ref{tabLevels} sums up the characterization of the operational situations we intend to implement in \textsc{Smaart}. 

\begin{table}
	\begin{center}
	\begin{tabular}{|c|l|}
\hline
1 & 
No activity on the part of the 
operator (no command sent to the 
UAVs)\\
\hline
2 & At least a command sent to a 
UAV in the last few minutes\\
\hline
3 & One alarm in the last few minutes\\
\hline
4 & Several alarms in the last few 
minutes\\
\hline
	\end{tabular}
	\end{center}
	\caption{Characterization of the four levels of mission workload in \textsc{Smaart}.}
	\label{tabLevels}
\end{table}

\begin{figure}
	\begin{center}
			\includegraphics[width=.8\textwidth]{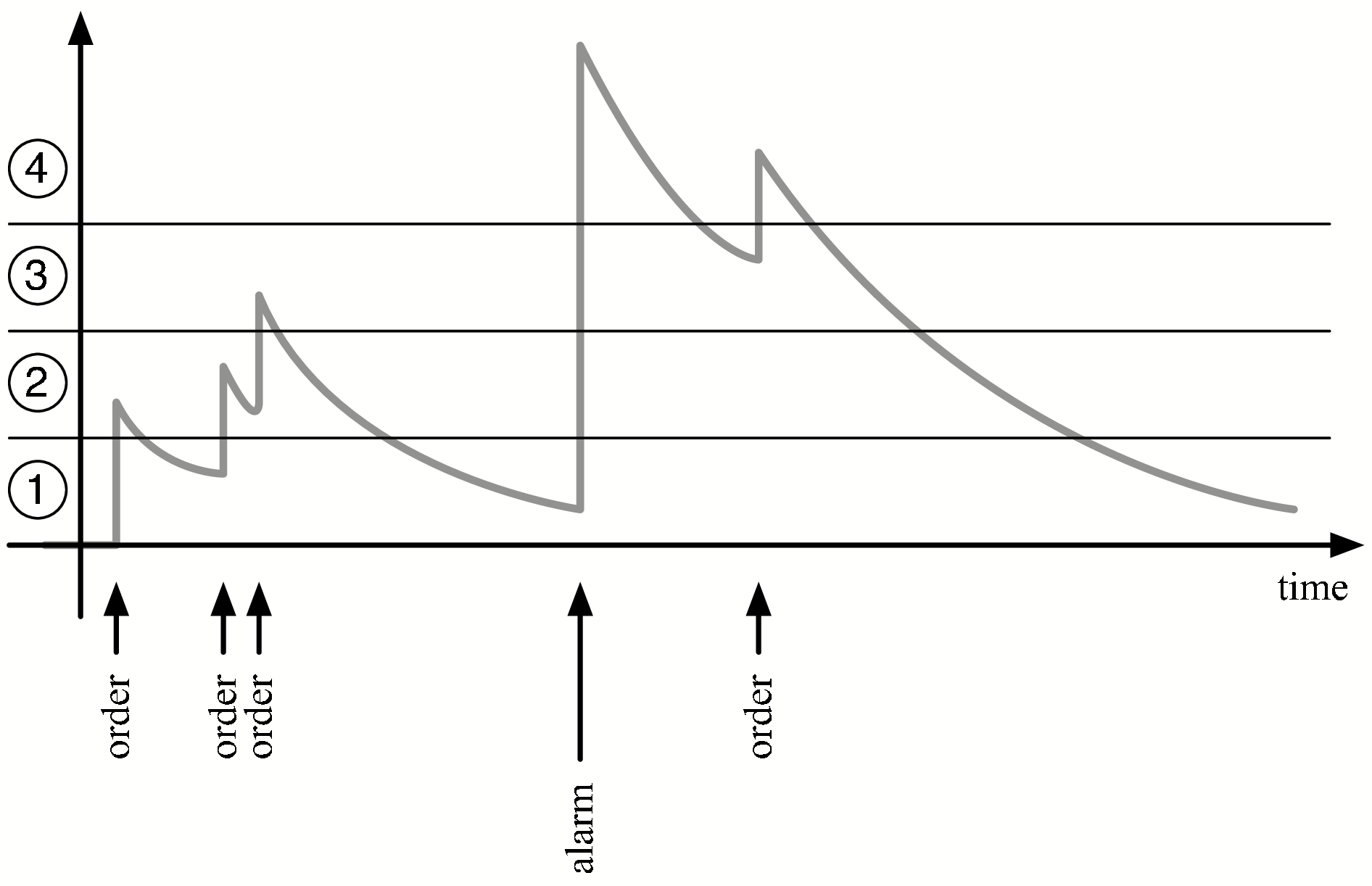}
	\end{center}
	\caption{Illustration of the computation of the four levels of mission workload.}
	\label{figLevels}
\end{figure}

Based on these criterions, the interaction manager is able to compute a discrete mission workload level at every moment: either (1) by storing every events (operator action toward UAVs or alarms) and matching with the criterions of table~\ref{tabLevels}, or (2) by updating a continuous   workload level by the combination of fixed additive values associated to alarms and orders with a discount temporal factor (see Figure~\ref{figLevels}). With the latter option, the continuous level is compared to pre-defined thresholds to obtain discrete levels.

\section{Conclusion \& Perspectives}

In the broad context of authority sharing, we have outlined how introducing more natural interaction in the design of the ground operator interface of UV systems should help in allowing a single operator to manage the complexity of his/her task. Introducing new modalities is one one of the \emph{means} in the realization of our vision of next-generation GOI. A more fundamental aspect resides in the \emph{interaction manager} which should help balance the workload of the operator between mission and interaction, notably by applying a multi-strategy approach to generation and interpretation.

We intend to apply these principles to the context of the \textsc{Smaart} prototype, and in this perspective, we have illustrated how to characterize the workload associated with a particular operational situation.

\small
\bibliographystyle{plain}
\bibliography{enstb,Bib-Article-HUMOUS-08,humous08biblio}

\begin{thebibliography}{10}

\bibitem{BC03}
A.~Bangerter and H.H. Clark.
\newblock Navigating joint projects with dialogue.
\newblock {\em Cognitive Science}, 27:195--225, 2003.

\bibitem{CWME03}
G.~Cai, H.~Wang, and A.~M. Mac~Eachren.
\newblock Communicating vague spatial concepts in human-{GIS} interactions: A
  collaborative dialogue approach.
\newblock {\em Spatial Information Theory}, pages 287---300, 2003.

\bibitem{CHPJ06}
J.Y.C. Chen, E.C. Haas, K.~Pillalamarri, and C.N. Jacobson.
\newblock Human robot interface: Issues in operator performance, interface
  design, and technologies.
\newblock Technical Report ARL-TR-3834, Army Research Laboratory (ARL),
  Aberdeen, July 2006.

\bibitem{Cla96}
H.~H. Clark.
\newblock {\em Using language}.
\newblock Cambridge University Press, Cambridge, UK, 1996.

\bibitem{MTNO90}
P.~Cohen, H.~Levesque, J.~Nunes, and S.~Oviatt.
\newblock Task-oriented dialogue as a consequence of joint activity.
\newblock In {\em Proceedings of PRICAI-90}, pages 203--208, 1990.

\bibitem{CBMM07}
M.L. Cummings, S.~Bruni, S.~Mercier, and P.J. Mitchell.
\newblock Automation architecture for single operator, multiple {UAV} command
  and control.
\newblock {\em The International Command and Control Journal}, 1(2):1--24,
  2007.

\bibitem{DW01}
S.~Dixon and C.~Wickens.
\newblock Control of {multiple-UAVs} : A workload analysis.
\newblock In {\em Proceedings of the 12th International Symposium on Aviation
  Psychology}, 2001.

\bibitem{End95}
M.R. Endsley.
\newblock Toward a theory of situation awareness in dynamic systems.
\newblock {\em Human Factors}, 37(1):32--64, 1995.

\bibitem{GNBWSC02}
D.V. Gunn, W.T. Nelson, R.S. Bolia, J.S. Warm, D.A. Schumsky, and K.J.
  Corcoran.
\newblock Target acquisition with {UAVs}: Vigilance displays and advanced
  cueing interfaces.
\newblock In {\em Proceedings of the Human Factors and Ergonomics Society 46th
  Annual Meeting}, pages 1541--1545, 2002.

\bibitem{Joh03}
C.~Johnson.
\newblock Inverting the control ratio : Human control of large, autonomous
  teams.
\newblock In {\em Proceedings of AAMAS'03 Workshop on Humans and Multi-Agent
  Systems}, 2003.

\bibitem{Kim93}
W.S. Kim.
\newblock Graphical operator interface for space telerobotics.
\newblock In {\em Proceedings of the IEEE International Conference on Robotics
  and Automation}, pages 761--768, 1993.

\bibitem{LC07}
F.~Legras and G.~Coppin.
\newblock Autonomy spectrum for a multiple {UAVs} system.
\newblock In {\em COGIS' 07 - COgnitive systems with Interactive Sensors},
  2007.

\bibitem{legras2008experiments}
Fran{\c c}ois Legras.
\newblock Experiments in human operation of a swarm of {UAVs}.
\newblock In {\em Proceedings of the first conference on Humans Operating
  Unmanned Systems (HUMOUS'08)}, Brest, France, 3-4 Sept. 2008.

\bibitem{LGCP02}
O.~Lemon, A.~Gruenstein, L.~Cavedon, and S.~Peters.
\newblock Collaborative dialogue for controlling autonomous systems.
\newblock In {\em Proceedings of AAAI Fall Symposium}, 2002.

\bibitem{MT03}
B.~Martinovski and D.~Traum.
\newblock Breakdown in human-machine interaction: the error is the clue.
\newblock In {\em Proceedings of the ISCA tutorial and research workshop on
  Error handling in dialogue systems}, pages 11--16, 2003.

\bibitem{MGKH01}
M.~Mouloua, Gilson, J.~R., Kring, and P.A. Hancock.
\newblock Workload, situation awareness, and teaming issues for {UAV/UCAV}
  operations.
\newblock In {\em Proceedings of the Human Factors and Ergonomics Society},
  volume~45, pages 162--165, 2001.

\bibitem{PSW00}
R.~Parasuraman, T.B. Sheridan, and C.D. Wickens.
\newblock A model for types and levels of human interaction with automation.
\newblock {\em IEEE Transactions on Systems, Man, and Cybernetics - Part A:
  Systems and Humans}, 30:286--297, 2000.

\bibitem{PG04}
M.~J. Pickering and S.~Garrod.
\newblock Toward a mechanistic psychology of dialogue.
\newblock {\em Behavioral and Brain Sciences}, 27:169--225, 2004.

\bibitem{SG06}
S.~Saget and M.~Guyomard.
\newblock Goal-oriented dialog as a collaborative subordinated activity
  involving collaborative acceptance.
\newblock In {\em Proceedings of Brandial'06}, pages 131--138, University of
  Potsdam, Germany, 2006.

\bibitem{SLC08}
S.~Saget, F.~Legras, and G.~Coppin.
\newblock Collaborative model of interaction and unmanned vehicle systems'
  interface.
\newblock In {\em HCP workshop on "Supervisory Control in Critical Systems
  Management", 3rd International Conference on Human Centered Processes
  (HCP-2008)}, Delft, The Netherlands, 2008.

\bibitem{SMW04}
D.H. Sonnenwald, K.L. Maglaughlin, and M.C. Whitton.
\newblock Designing to support situation awareness across distances: an example
  from a scientific collaboratory.
\newblock {\em Information Processing \& Management}, 8(6):989--1011, 2004.

\bibitem{Tra94}
D.~Traum.
\newblock {\em A Computational Theory of Grounding in Natural Language
  Conversation}.
\newblock PhD thesis, Computer Science Deptartment, University of Rochester,
  1994.

\bibitem{ullman2007oo-oo-oo}
David~G. Ullman.
\newblock ``{OO-OO-OO!}'' the sound of a broken {OODA} loop.
\newblock {\em Crosstalk}, April 2007.

\bibitem{WHKW01}
J.~Wesp, R.and~Hesse, D.~Keutmann, and K.~Wheaton.
\newblock Gestures maintain spatial imagery.
\newblock {\em American Journal of Psychology}, 114(1):591--600, 2001.

\bibitem{WDCB05}
D.T. Williamson, M.H Draper, G.L. Calhoun, and T.P. Barry.
\newblock Commercial speech recognition technology in the military domain:
  Results of two recent research efforts.
\newblock {\em International Journal of Speech Technology}, 8(1):9--16, 2005.

\end{thebibliography}

\end{document}